# A survey on the development status and application prospects of knowledge graph in smart grids

Jian Wang, Xi Wang, Chaoqun Ma, Lei Kou

## Abstract

With the advent of the electric power big data era, semantic interoperability and interconnection of power data have received extensive attention. Knowledge graph technology is a new method describing the complex relationships between concepts and entities in the objective world, which is widely concerned because of its robust knowledge inference ability. Especially with the proliferation of measurement devices and exponential growth of electric power data empowers, electric power knowledge graph provides new opportunities to solve the contradictions between the massive power resources and the continuously increasing demands for intelligent applications. In an attempt to fulfil the potential of knowledge graph and deal with the various challenges faced, as well as to obtain insights to achieve business applications of smart grids, this work first presents a holistic study of knowledge-driven intelligent application integration. Specifically, a detailed overview of electric power knowledge mining is provided. Then, the overview of the knowledge graph in smart grids is introduced. Moreover, the architecture of the big knowledge graph platform for smart grids and critical technologies are described. Furthermore, this paper comprehensively elaborates on the application prospects leveraged by knowledge graph oriented to smart grids, power consumer service, decision-making in dispatching, and operation and maintenance of power equipment. Finally, issues and challenges are summarised.

## 1 INTRODUCTION

The conventional electric power systems can no longer meet the huge requirements of information age due to the continuous improvement of the economy, increasing demands for electric power, public awareness of green energy, large-scale incorporation of intermittent renewable energy penetration, and wide application of information and communication technologies [1]. To solve the energy crisis and environmental pollution, various renewable energy technologies have been developed rapidly [2-4]. However, due to the randomness and intermittency of renewable energy sources, the large-scale applications of various renewable energy technologies have adversely placed great pressure on the safety and reliability of traditional power systems. Hence, smart grids (SGs) play a historic role in safely plunging renewable energy resources into the highly controllable grid to supplement the power supply ensured by smart communications, sensors and measurement devices [5].

In contrast with the traditional power grids, SGs belong to the most significant evolutionary developments in panoramic real-time systems because they integrate the actions of all users connected to it—generators, consumers and those that do both—in order to efficiently deliver sustainable, economic and secure electricity supplies [6]. At the same time, the massive data generated in power generation, power transformation and transmission, distribution network, and electricity consumption is combined to form electric power big data [7, 8], which has great potential to support the safety and stability operation of power grids, high-quality business services, and diverse decision-making [9]. However, it also leads to great challenges in big data processing analysis and knowledge mining. In order to meet the needs of electric power data management, power consumer-oriented services, and SGs business development, a semantic knowledge integration platform should be proposed to obtain advanced SGs scenario applications and business services, merging multi-source heterogeneous information from multiple SGs scenarios. The knowledge graph (KG) [10] provides us an effective solution to achieve this goal, which is crucial for facilitating the intelligentization of power grids as well. Taking power equipment management as an example, the KG can improve planning, traceability, production scheduling, order execution, and supplier management of power equipment [11].

A KG is a multi-relational graph composed of entities (vertices) and relations (different types of edges). Each edge is represented as a triple of the SPO (subject, predicate, object) form, also called a fact, indicating that two entities are connected by a specific relation, for example, (Eric Emerson Schmidt, is the president of, Google) [12]. KG technology, a novel means to extract semantic knowledge from structured web data, text, and images, has already had many applications, for example, semantic search, information integration, knowledge management, question answering, and decision-making. Over the last decade, the knowledge graphs (KGs) have been widely applied in various modern domains such as the medical domain, electronic commerce, and finance, to virtually merge knowledge from multi-source data into knowledge bases [13-16]. Meanwhile, there are many successful applications of large-scale knowledge systems, including Freebase [17], DBpedia [18], YAGO [19], NELL [20], and Wikidata [21]. Therefore, it has been proved that KGs describe the real world with the interpretable graph structure, store structured relational facts of concrete entities and abstract concepts in the real world from a novel semantic perspective. The structured relational facts could be either automatically extracted from enormous plaintexts and structured web data, or manually annotated by human experts [22]. In such a case, it is believed that knowledge graph can also be widely applied in power systems and contribute to beneficial and fruitful results, especially the utilisation and management of the electric power big data. In this regard, compared with the studies of KGs in other domains, the KG for SGs (KG4SGs) is a huge semantic network that combines the encyclopaedic knowledge, business data, and power industry specifications according to the power business architectures, electric power knowledge architectures, and power industry standards. To this end, KG4SGs offers a new data integration paradigm that merges large-scale data processing with robust semantic technologies, making first steps towards the next generation of electric power artificial intelligence [11].

The advancements of power systems and Energy Internet have opened the door to new characteristics with far-reaching impacts. For example, energy and information are intertwined; new energy continues to account for at a high rate; more renewable energy sources, electric vehicles, and distributed generators are integrated into the network; power trading and market are playing an increasingly important role. Artificial intelligence (AI), one of the most disruptive technologies in the world, demonstrates strong processing ability in computational intelligence, perceptual intelligence, and cognitive intelligence. The application of AI in power systems will promote safe operation of the power systems, change the service modes of businesses, and provide a significant means to support a new generation of power systems. In other words, electric power artificial intelligence will become an important strategic direction as well as an inevitable solution for the future development of power grids. Moreover, electrical power science has formed rich knowledge systems in long-term operation. However, highly abstracted knowledge data often exists in text form. Hence, it is essential and indispensable to explore how to use big data technology, deep computing technology, and natural language processing (NLP) technology to extract power professional vocabulary, construct electric power professional lexicons, and then build KG4SGs to achieve self-understanding, self-decision-making and autonomous behaviour of power system operation. Ultimately, KG4SGs not only effectively supports grid dispatching, transmission inspection, distribution, electricity, customer service, and other services, but promotes the development of power grids toward human-like intelligence.

Up to this moment, the smart grids and the KG have been usually reported separately in [5, 23, 24] and [12, 25, 26]. However, specific research on KG towards SGs is still quite limited. To the best of our knowledge, Tang et al. [11] merged existing multi-source heterogeneous data of power equipment to construct a KG aiming at power asset management. Chun et al. [27] proposed an energy KG to provide the semantic integration of various energy services. Since the conventional knowledge base for power information acquisition system were unable to obtain efficient decision-making, Feng et al. devised a solution to carry out operation and maintenance of telecommunication information acquisition by using KG [28]. In order to further realise the potential of KG in SGs, this paper makes the first attempt to systematically look into the KG issues in SGs. The objective of this paper is to estimate with what degree KG technology is applied in SGs and to identify its further potential. To be specific, the more recent research on power text mining, electric power knowledge ontology learning and social media analysis are reviewed to provide the extent to which existing electric power knowledge mining is insufficient and KG4SGs technology is only in its infancy. Moreover, the definition, constructing method, the framework of the KG technology is presented and the advantages and demand analysis of smart grids for knowledge graph are discussed. On the other hand, a big knowledge platform for SGs is devised to provide a better understanding of the full potential and demonstrate how KG4SGs shapes the modern electricity grid applications. This paper highlights the key research issues in KG4SGs, including its concept, construction methods, characteristics, and application prospects. It also intends to identify research activities, trends, issues, and challenges. The more researchers know about the KG4SGs, the better they will understand its value. In this case, they will be more willing to recognise that KG4SGs is essential for the interdisciplinary analysis of electric power and AI.

The remaining sections are organised as follows. A review of electric power knowledge mining is conducted in Section 2. The overview of the KG in smart grids are introduced in Section 3. The architecture of KG4SGs and key technologies is illustrated in Section 4. The prospects of three application scenarios in SGs are elaborated on in Section 5, including power consumer service, decision-making in dispatching, and operation and maintenance of power equipment.

Section [6] extensively summarises the related issues and challenges. Finally, a conclusion is made in Section [7].

# 2 RESEARCH STATUS OF ELECTRIC POWER KNOWLEDGE MINING

Various previous research on knowledge mining related to the electric power KG construction has been undertaken by both corporations and academia, which globally uses ontology learning, text mining, semantic understanding, and social media analysis technologies to excavate key information, achieve various binary relations expression and interoperability, and then support the SGs knowledge engineering formulation and the intelligentization of the electric power network. In this section, a brief overview of existing electric power knowledge mining is provided. The objective of the review is to present the methods and developmental limitations in the existing studies of electric power knowledge mining and lay a foundation for the research on KG.

## 2.1 Ontology learning

Ontology refers to the formal, explicit, and shared conceptual explanation of real world. In other words, an ontology is a formal representation of a set of concepts and their relationships in a particular domain, which is suitable for semantic information representation and inference [[29], [30]]. Meanwhile, a well-constructed ontology can facilitate machine processable definitions and help develop the knowledge-based information search and management systems more effectively and efficiently [[31]]. Ontology-based applications in SGs can be specifically classified into two aspects, that is, power substation and energy system.

Feng et al. investigated the transformer temperature modelling based on ontology and agent to conduct thermal analysis [[32]]. Liao et al. [[33]] combined association rules with ontology to analyse substation alarm information, which transformed the structured and unstructured data into extensible markup language (XML) data that allowed logical reasoning and numerical calculation. For power system asset management, Yan et al. [[34]] introduced a kernel-based consensus clustering algorithm embedded with domain ontology to improve document repository of power substations. Based on the ontology and semantic framework, Wang et al. compiled the power equipment ontology dictionary to extract defect components and attributes of power equipment [[35]].

In terms of ontology-based applications in energy system, Gaha et al. [[36]] used common information model (CIM) ontology to solve the problem of semantic conflict in electric power systems. A basic universal method of electric power system knowledge expression based on the ontology and semantic web was proposed in [[37]], which adapted to various knowledge expression requirements of electric power big data. With the advent of electric power big data era, unstructured document extraction has become a new important issue for all energy utilities. Kumaravel et al. [[38]] utilised a multi-domain layered ontology model to excavate the key information from unstructured documents and build thermal power plant ontology, which could be extended to any industry by integrating appropriate domain ontology with the domains. To obtain the general description of complex systems, Jirkovský et al. [[39]] used the semantic sensor network ontology to describe the cyber-physical system (CPS) from the perspective of component.

To detect security intrusion or attacks of the power Internet of Things and cloud (IoT-Cloud), Choi et al. [40] devised an ontology-based security context reasoning method to improve the security service. To orchestrate the utility business processes, Ravikumar et al. [41] developed the CIM-based process ontology to model end-to-end process operations of power utilities. In renewable energy storage systems, Maffei et al. [42] added an ontology subsystem to achieve the unified data formatting and provided semantics to the middleware.

To sum up, there are various studies of ontology applied in power substation as well as energy systems, and ontology learning is playing a prevalent and essential role in developing electric power knowledge models and expression. However, existing ontology learning methods are mainly based on single domain and keywords integration, thus lacking in-depth semantic analysis, which is insufficient and ineffective for the expanding knowledge base of SGs and the imperious demands from business services. Particularly, the construction of each sub-domain ontology towards grid scenarios is independent, and semantic heterogeneity exists in the whole electric power knowledge flow [38]. It is widely expected that many ontologies of different electric power domains in different languages will be developed. Meanwhile, there are great challenges in interoperability and interconnection during the creation and maintenance of the power ontology learning for SGs.

## 2.2 Power text data mining

Text mining, a part of data mining, can discover the underlying knowledge from textual data. In SGs, a large number of texts related to the operation and control of power grids are widely accumulated, including trouble and defect records, operating tickets, logs of operation and maintenance and so on. Text mining [43-45] plays a vital role in extracting critical information from electric power texts, such as equipment name, equipment type, location, the logical connection of equipment. Particularly, Chinese texts possess the general characteristics of obscure, ambiguous, and hardly segmenting [46].

Text data mining provides new and essential insights for asset management decision makers [47], condition assessment [48], deep analysis of power equipment defect [35, 49], and analysis of power customer appeals [50]. Xie et al. used HMM-based (hidden Markov model) text reprocessing to extract the key information from fault and defect elimination record texts to assess the operating condition of distribution transformer, combined with typical power-off tests and live line detecting results [48]. To identify the causes of transformer failure, Ravi et al. studied 393 terms and 103 documents and found that 'leak', 'lightning', 'animal', 'cable' and 'temperature' were the main causes [49]. However, these approaches lead to poor results since they failed to consider the semantic and contextual information and the characteristics of electric power knowledge. A few studies with focus on semantic analysis were presented in [35] and [50]. Wang et al. proposed a deep semantic scheme of defect mining based on 'Semantic Frame Slot Filling' (SF-SF) [35]. The unstructured information in historical defect texts was transformed into structured information, and defect components and attributes were extracted for further research. Sheng et al. [50] utilised a multi-algorithm-based model involved with Adaboost, Support Vector Machine (SVM), and Random Forest to obtain power customer appeals from 95598 voice-transcription texts for improving service quality.

Overall, text mining has been preliminarily applied in power systems, assisting in the early diagnosis of power equipment, power consumer appeals analysis as well as in capturing dispatching experience by extracting the key information from textual records such as power equipment asset failure and defect texts, power consumer complaints, and dispatching tickets [48]. However, how to parse the texts from a semantic perspective so that key information can be extracted for defect analysis is still an unresolved issue. Furthermore, there are significant differences in the algorithms applied in the multilingual text. For example, English text is found to be incompatible and untranslatable in Chinese text since the structure of Chinese characters is more complex [51]. In addition, studies of the power system domain pertaining to text mining from a semantic perspective are still rather limited.

## 2.3 Social media analysis

Social networking is the most popular online activity and 91% of netizens use social media regularly. Facebook, YouTube, and Twitter are the first, third, and tenth most-trafficked sites on the Internet [52], which are also considered as social sensors. The heterogeneous information with various formats from social sensors has been used in the real-time event extraction [53] such as disasters [54, 55], power outage [56], traffic, and severe weather events. Sakaki et al. [54] developed a probabilistic spatio-temporal model leveraged by Kalman filtering and particle filtering to detect events and estimate the location. Moreover, an earthquake reporting system was constructed and applied in Japan. Lee et al. [55] applied the Naive Bayes classifier to identify the tweets related to the blackout, however the error rate was relatively high.

Power system analysis is different from earthquakes and hurricanes, hence these methods cannot directly be applied to monitor power outages [57]. The integration of social media with power system analysis and operation is a relatively new topic, and the usefulness of social media in outage detection has been recognised by the power industry. Sun et al. [56] integrated the textual, temporal, and spatial information and conducted a detecting and locating method of power outages based on latent Dirichlet allocation. Bauman et al. [58] proposed a keywords relevant detecting method to identify power outages from the tweets for discovering emergency events, particularly focusing on 'small events' related to the general public. However, it should be noted that the success of applying the above methods to social media analysis depends on how well the domain experts could set the key parameters and the system inputs for those applications.

In this section, existing knowledge discovery domains of ontology, power text mining, and social media analysis are comprehensively surveyed. Since SGs knowledge is characterised by diversity, correlation, synergy, and hiddenness, previous research on these three issues is only superficial analysis and lacks deep semantic analysis, domain-specific knowledge model, and facts association for SGs scenarios.

    Diversity. There are many types of power equipment and several sub-domains in SGs, which can in turn produce massive diverse data and knowledge.

    Correlation. There exists association or dependency between an event and another event. This is because associated power knowledge widely exists in SGs, hence it is of great significance to excavate correlation knowledge during the process of knowledge reasoning.

Synergy. Decision-making often calls for comprehensive knowledge analysis because none of the knowledge can exist separately.

Hiddenness. A great deal of raw data and information in the power grids might be incomplete, noisy, fuzzy, and random. Although they have no practical significance themself, and the truly valuable knowledge is hidden behind these data and information, which needs to be discovered through certain means of knowledge discovery.

The challenges brought by electric power big data are not only the huge scale of data in a pure sense, but also the improvement of big data analysis technologies to meet the increasingly diverse requirements for personalised services and knowledge navigation. The next thing need to be considered is how to extract and analyse valuable knowledge from massive data, which is among the top concerns of research on big data. Therefore, it is essential to understand the concepts and terminology of power grids as well as the relationships between them, study semantic heterogeneity, establish the specific knowledge database for electric power big data towards situational awareness analysis and further provide personalised services for power users according to the experiences, roles, and tasks of users.

## 3 THE OVERVIEW OF THE KNOWLEDGE GRAPH IN SMART GRIDS

### 3.1 What is the knowledge graph

A KG [59] is a network of all kinds of entities, relations, and attributes related to a specific domain or topic, which provides a programmatic way to model a specific real-world domain with the assistance of subject–matter experts, data interlinking, and machine learning algorithms [60]. A KG is typically built on top of the existing databases to link all data together at web-scale combining both structured and unstructured data, including objects, abstract concepts, numbers, and documents. It is a directed edge-labelled graph, whose nodes represent entities and properties of interest and whose edges represent relations between these entities.

Each link of two nodes is usually represented as a triple of SPO, which can be recognised as a piece of knowledge. Moreover, two endpoints allow the multiple links because there are more than one relationship between two entities. For example, (oil tank, is a part of, the transformer), (the operating condition of oil tank, seriously influences, the transformer). Formally, a directed graph can be modelled as a tuple     [61-63], where     is a set of entities (i.e. subjects or objects) and    is a set of edges that map all the relation types, where     . Each possible triple and binary random variable can be model as     and    . Then, all the triples in     can be represented as a three-way array    , whose entries are set such that

(1)

Figure 1 illustrates the overall architecture of KG technology [64]. The part in the dotted box is the process of constructing and updating KG. The KG construction is to extract knowledge elements (i.e. facts) from raw data by several automatic or semi-automatic techniques, and then extracted entities (concepts), attributes, and relationships between entities are stored in to knowledge base.

The whole process is dynamic and iterative and relevant construction/update procedures consist of four steps: (1) information extraction [65]: extracting entity (concept) attributes and their interrelations from various data sources and forming ontology-based knowledge representation. The key technologies involved: entity extraction [66], relation extraction [67], and attribute extraction; (2) knowledge representation: representing entities and relations in KG in low-dimensional semantic space. The typical models for knowledge representation learning including linear model, neural model, translation model and other models [12, 26, 68]; (3) knowledge fusion: merging new knowledge, and eliminating contradictions and ambiguities [69]. The key technologies involved: entity linking, entity alignment, entity disambiguation, and knowledge reasoning [70]; (4) knowledge processing: the integrated knowledge will only be merged into knowledge base after it has undergone quality evaluation or manual judgment. The key technologies involved: quality evaluation, knowledge graph refinement, knowledge graph completion, and knowledge graph correction [25].

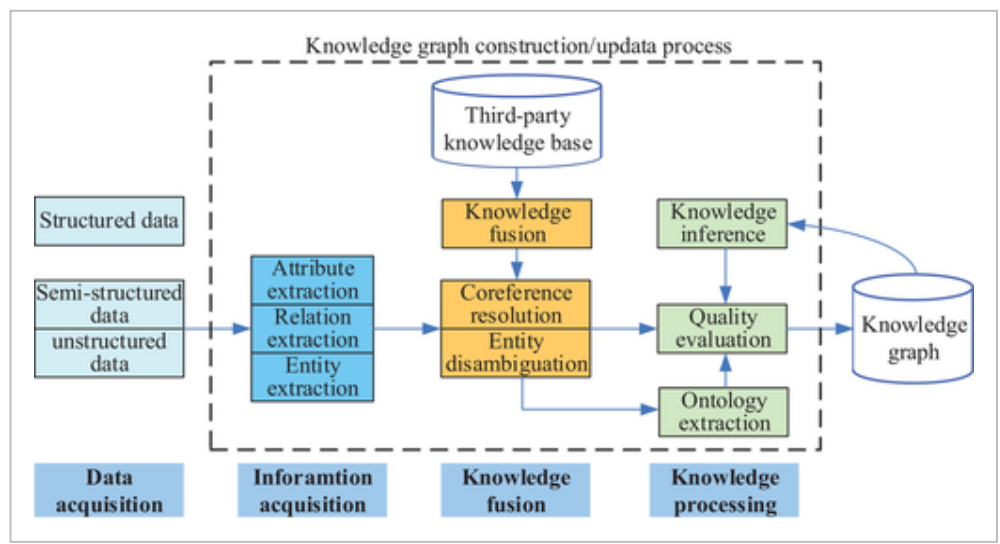

**FIGURE 1**

Open in figure viewer | PowerPoint

The framework of knowledge graph

## 3.2 Why is the knowledge graph

As mentioned above, the KG refers to a way of storing, interlocking and organising the datasets and knowledge, which allows people and machines to better tag the content and properly relations into the connected datasets. To the best of our knowledge, KG is becoming ubiquitous, powering everything from recommendation engines and enhanced query to NLP and conversational applications. There are many reasons for making use of KG, which are as follows:

Breaking down all those data silos. The KG makes the machines understand real-world context with a flexible data layer, which integrates the real-word facts, events and concepts from the perspective of a so-called 'unified view'.

Finding information faster. Fundamentally, a KG is a graph database which stores the knowledge and information in a graphical format, which means the relationships between any data points can be calculated far more quickly and with less compute power overheads.

Making better decisions. The more enriched and in-depth search results can be captured with the help of networks of 'things' and facts.

Uncovering a whole lot of hidden insights. The existing AI technologies are extreme black-box models, which lead to the facts that the operators are unaware of internal knowledge flow and how the black-box algorithms make decisions. The KGs enriched with entities, relations and concepts help with AI explainability.

## 3.3 The demand analysis of smart grids for knowledge graph

As smart grids thrive rapidly, massive advanced metering infrastructure and sensors are deployed in electric power systems. At the same time, an unprecedented amount of multi-source heterogeneous big data is accumulated. It is an urgent problem to construct a full-service unified data centre for intelligently analysing and managing large volume of data [71]. To the best of our knowledge, the KGs provide a flexible way to establish semantic connections and obtain unified semantic-level data service, known as an ontology. The KG4SG is a huge semantic network that merges the multi-source heterogeneous big data, which leverages a new data integration paradigm that is applicable to the next generation of electric power artificial intelligence [11]. Moreover, KG4SG can be widely applied in equipment failure analysis, consumer service, fraud data analysis and other fields.

KGs give electric power AI applications intelligence. For example, it can provide relevant facts and contextualised answers to your specific questions. In addition, electrical power AI benefits from KGs for their enhanced query and search, and then KG4SG help AI discover hidden facts and relationships through inferences in the integrated content that operators would otherwise be unable to catch on a large scale.

## 4 BIG KNOWLEDGE GRAPH PLATFORM FOR SGs

In order to provide a better understanding of the business applications and the development of SGs, an integrated enterprise-level information integration platform should be proposed to realise the smooth flow of electric power big data and information sharing between power consumers and electric power utilities. A well-integrated big knowledge graph (BKG) platform can organise, construct, manage and make use of large-scale KGs, which not only ensures efficient operation of power systems, but also benefits all the key stakeholders (e.g. power utilities, operators, consumers) [72]. In such a case, a framework of KG for SGs is devised as shown in Figure 2 based on [91-93]. The whole KG4SGs consists of four layers [94], namely, data acquisition layer, knowledge graph layer, knowledge computation and management layer, and intelligent application layer, which follows a hierarchical pattern according to electricity knowledge flow. This section focuses on the first three layers. The intelligent application layer will be elaborated on particularly in Section 5.

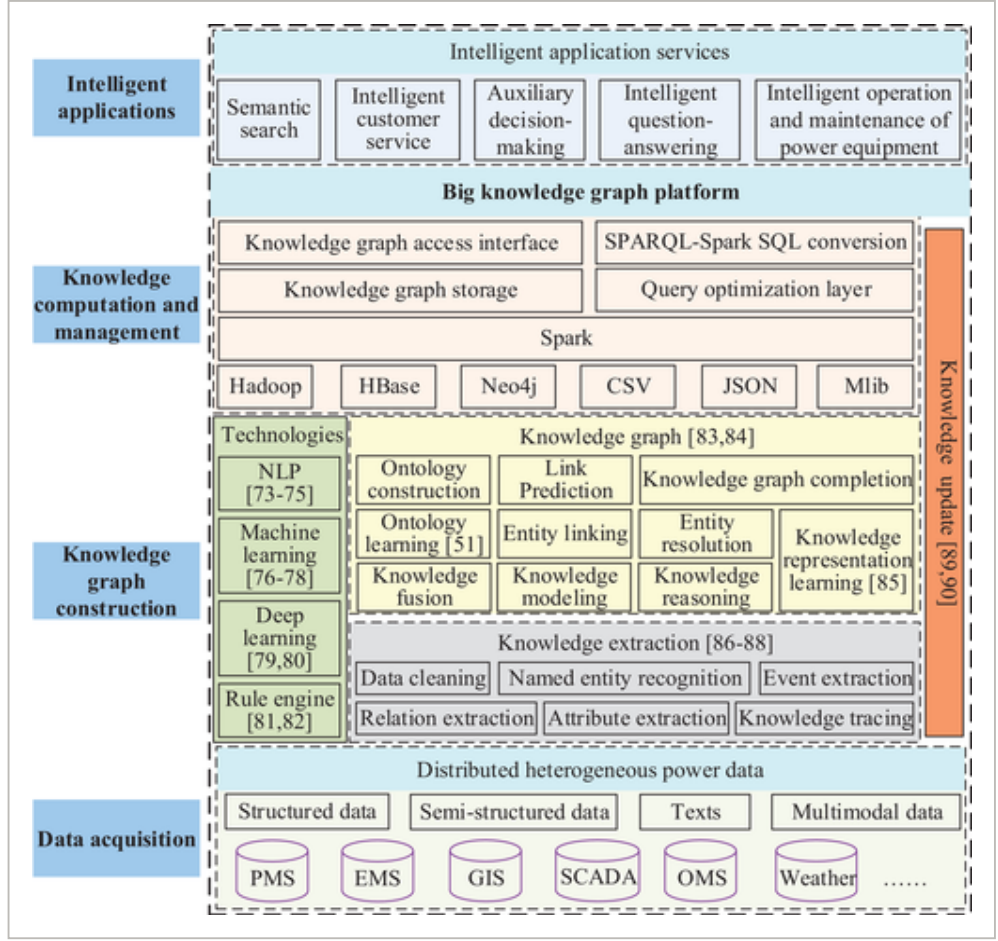

**FIGURE 2**

The framework of knowledge graph towards smart grids

The high-level design of the BKG platform for SGs is developed in Figure 2, which can provide a variety of functionalities aiming at SGs businesses and intelligent applications. That is, distributed power heterogeneous data generated from the operation and maintenance of the complex power grids, power equipment, and power consumers can be transformed into the electricity knowledge via knowledge extraction technologies. So that the entities (e.g. real objects, line failures, equipment failures, location of line failures, and power equipment), attributes (e.g. failure characteristics, English names of failure and types of failure), relations (e.g. the causes of failure, treatment methods, selection and application methods), and events (e.g. power outage, thunder, severe weather) related to electricity knowledge are excavated from the digital data, images, textual defect records and so on by knowledge processing. Then, various types of relations between excavated entities can be integrated into a network (i.e. KG), in which an entity corresponds to a vertex and a relation corresponds to a directed edge. Moreover, each links of the two nodes related to the electricity knowledge is represented as SPO form, which is beneficial for the electric knowledge engineering to facilitate facts association. To store massive entities, attributes, relations and events of SGs and build more intelligent machine learning algorithms to enable knowledge flow capability and efficiency, the big data platform is designed for the implement of graph data models (HBase, Neo4j), graph algorithms, database engines (Spark, Hadoop) and database interface. In particular, the big knowledge platform also supports intelligent services towards diverse application scenarios, for example, semantic search, intelligent power consumer service system, auxiliary decision-making, intelligent operation and maintenance of power equipment.

## 4.1 Data acquisition layer

Considering the fact that electric power big data is collected from disparate data platforms by different monitoring infrastructures, the objective of data acquisition layer is to integrate all the data connected to the SGs into the BKG platform, including online open websites, third-party integrated service platforms, electric power knowledge databases and so on [95]. According to the different data sources, the data in SGs can be divided into two categories. One refers to internal data of power systems, which is accumulated from production management system (PMS), energy management system (EMS), outrage management system (OMS), asset management system (AMS), condition-monitoring management system (CMS), etc. The other is external data, which consists of the meteorological information system (MIS), geographic information system (GIS), public service sector and internet. All these collected sub-domain hybrid corpora contain a large amount of information, which supports data and knowledge mining, the excavation of values, novel knowledge-driven algorithms and allows improvisation in existing operation and planning practices. Therefore, the effective and efficient integration of big data analysis are crucial for all aspects of the whole electric power knowledge chain, such as suppliers, operators, consumers, and regulators. [96].

### 4.1.1 Internal data

A. Trouble and failure records

In the operation of power grids and power asset management, inspectors record trouble tickets in text and electronic forms. These unstructured data not only reflects the historical trend of the operating condition of power equipment and infrastructure, but also contains rich latent fault information [97]. However, existing methods cannot exploit the trouble and failure records efficiently because of the complexity of text semantics and structures. For example, State Grid Corporation of China has stored large quantities of defect records in the PMS as shown in Table 1.

TABLE 1. Defect records information of power equipment

| Records type | Defect information |
| --- | --- |
| Equipment nameplate | Equipment number, name, equipment type, commissioning time, commission age, manufacturer number, etc. |
| Manufacturer | Manufacturer number, manufacturer name, production date, etc. |
| Equipment defect | Equipment number, component/part, occurrence time, defect description, technical reason, responsibility for defect, defect treatment, etc. |

B. Service tickets and monitoring data

There are large quantities of power consumer service tickets in the long-term power transactions and services. Meanwhile, customer satisfaction continues to be among the top concerns of power utilities. Electricity consumption behaviour analysis and new hot spots of consumer attention extracted from service tickets are attracting the interest of many researchers. As for power utilities, power consumer service system can provide active proactive perception and service. For example, it intelligently reminds the users if they need to query the content that they often pay attention to and pushes attention notifications. In addition, various wide-area monitoring systems like AMI, PMS, and electric power marketing system continuously generate massive amounts of operation data, which concerns the safe operation and reliability of power equipment assets, power grids, and utilities.

C. Authoritative standards and guides

Many official organisations and electric power companies have set a series of standards, covering traditional electricity systems, new SGs, and other related discipline standards.

Research institutions comprise IEC, IEEE, CIGRE, EPRI, W3C, ISO, the State Grid Corporation of China and so on, in which there are abundant information and the authoritative knowledge of many experts.

D. Domain expert knowledge

Abundant valuable knowledge and experience have been generated during the operation, maintenance, marketing of SGs in the long-term production of electric power. For example, an experienced dispatcher can accurately judge the safety margin of the power system operation and a senior maintenance engineer can determine whether the transformer is operating well by listening to the sound. Effective management of these intangible knowledge assets is of great practical significance and economic value.

### 4.1.2 Online open information

A. Social media

With the advent of the digital information era, apps such as Facebook and Twitter enable power consumers to engage in the operation and economic dispatch of SGs. The above discussions in Section **2** show that the data from social sensors can be used to identify the location and extent of an outage without other measurement and communication instruments [**56**].

B. Open professional literature in electric power field

On the Internet, there is much publicly available literature. Some academic literature search engines have provided professional retrieval of keywords based on NLP technology now. However, these search engines are mainly based on traditional regular matching techniques as they have no professional thesaurus and corpora. At present, the KGs led by Google are in full swing and have already been applied in the Internet and medical fields.

C. Power information websites

There are some open professional websites recording valuable information as shown in Table 2. Although the documents of these restricted topics are relatively sparse and their quality levels are varied, they can provide the auxiliary foundation for information retrieval and public opinion monitoring.

**TABLE 2.** Power information websites

| Website name | Website address |
| --- | --- |
| Edison electric institute | https://www.eei.org/pages/default.aspx |
| General energy online | http://www.genergyonline.com/ |
| Energy central | https://www.energycentral.com/ |
| United States department of energy | http://www.doe.gov/ |
| Electric net | https://www.electricnet.com/ |
| Electric power research institute | http://loadshape.epri.com/ |
| Electric power supply association | https://epsa.org/ |
| International electrotechnical commission | https://www.iec.ch/ |
| The institute of engineering and technology | https://www.theiet.org/ |
| International council on large electric systems | https://www.cigre.org/ |
| Power technology | https://www.power-technology.com/projects/ |
| Joint information systems committee | http://www.intute.ac.uk/sciences/cgi-bin/browse.pl?id=25667 |

## 4.2 Knowledge graph construction

Similar to the construction of conventional domain KG, the creation of the electric power KG adopts the bottom-up pattern, during which the key entities, relations, and attributes derived from large-scale heterogeneous power data are progressively processed, linked and added into a knowledge base in view for the actual demands in SGs scenarios [98]. The framework of electric power KG construction consists of three layers: knowledge acquisition layer, sub-domain knowledge graph layer, and electric power knowledge graph layer (Figure 3).

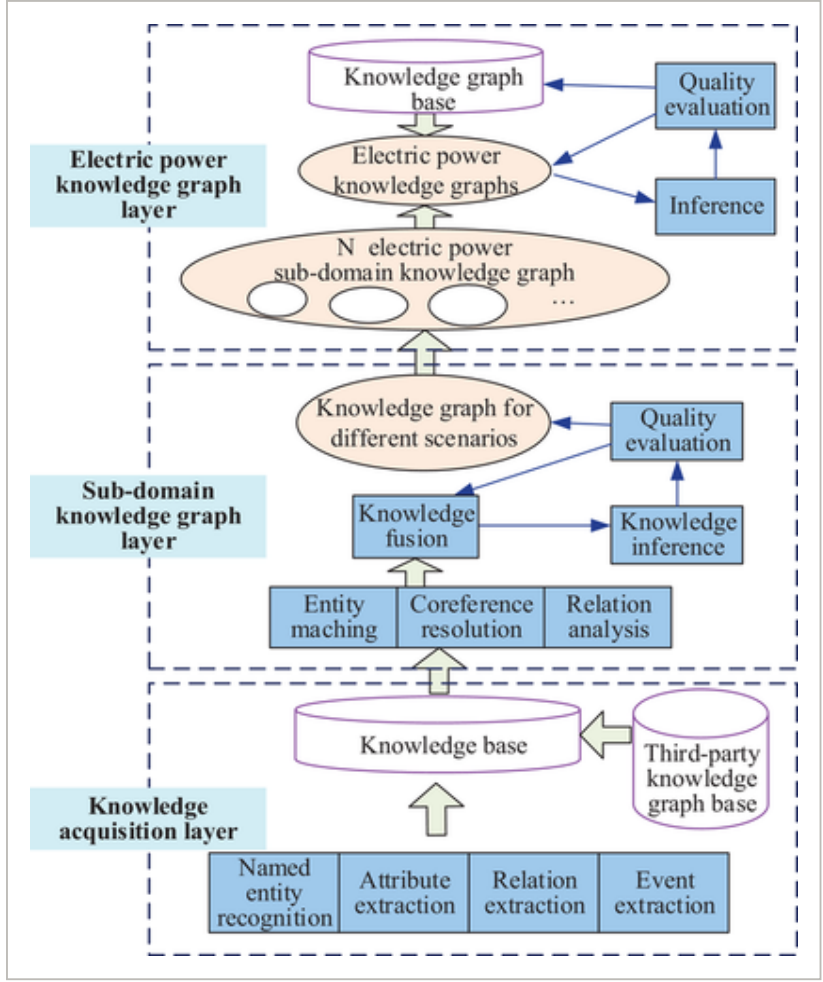

**FIGURE 3**

Framework of electric power knowledge graph [98]

### 4.2.1 Knowledge acquisition layer

In knowledge acquisition layer, numerous entities, entity attributes, and relations between entities from the distributed heterogeneous power resources are achieved by named entity recognition (NER), attribute extraction technology, relation extraction, and event extraction [67], and electric power knowledge base is established to store the electric power knowledge. Meanwhile, a third-party domain-specific expertise knowledge base built by experts can also be integrated into the constructed electric power knowledge base. Figure 4 describes an example of extracting entities from a defect record of transformer (oil storage cabinet of #1 main transformer has oil leakage, and the speed is 5 drop per minute) [35].

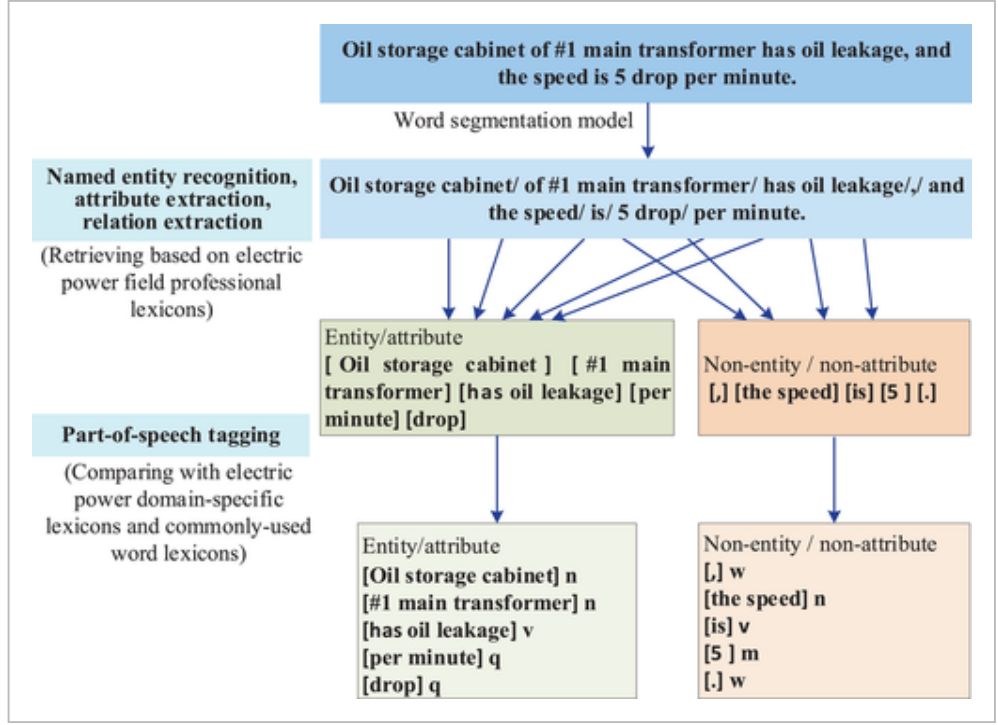

**FIGURE 4**

An example of entity/attribute extraction [99]

### 4.2.2 Sub-domain knowledge graph layer

In sub-domain KG layer, KGs of different business scenarios or power domains in SGs are constructed, such as power customer service system, decision-making in dispatching, operation and maintenance of electric equipment. For each sub-domain KG, it merely corresponds with one domain-specific demand. Specifically, entity matching, co-reference resolution, and relation analysis are adopted to eliminate the disambiguation and errors and reduce the redundancy of concepts and entities. To obtain a well-networked and structured electric power topic KG, the processes of knowledge fusion [69] (e.g. entity linking, and entity resolution) and knowledge inference are subsequently carried out to adjust and modify the obtained results. Since new knowledge and the results of knowledge mining might be incomplete and error-prone, quality assessment is particularly essential to discard knowledge with low confidence before adding new or extracted knowledge into an existing sub-domain graph.

A reflection on KG of defect records is depicted in Figure 5(a), which illustrates the entities and attributes of 'main transformer', 'oil storage cabinet', 'oil leakage', and visual relations between them. It is obvious that 'oil storage cabinet is a component of main transformer', 'oil storage cabinet has oil leakage', and 'the speed of oil leakage is 5 drop per minute'. The information of the transformer can be propagated on this graph. Figure 5(b) displays another transformer defect record on KG, oil conservator of #1 main transformer body has serious oil leakage.

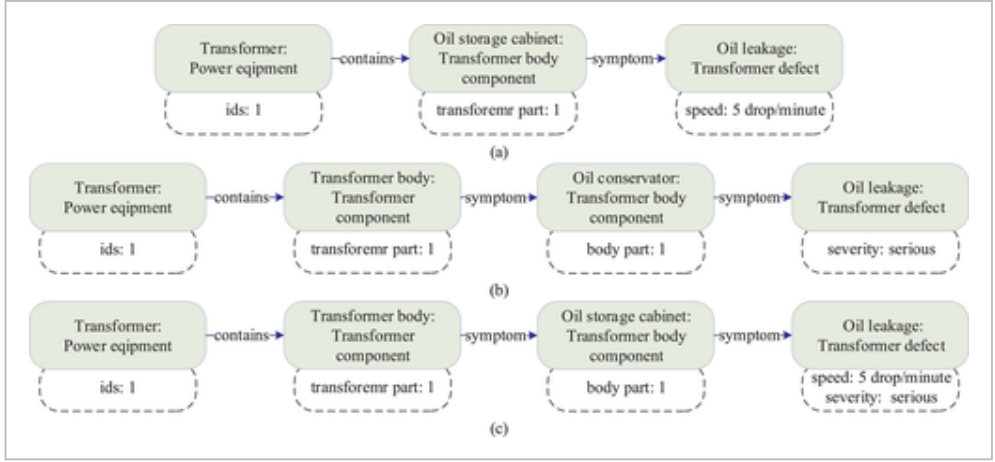

**FIGURE 5**

The reflection on knowledge graph of defect records. (a) Oil storage cabinet of #1 main transformer has oil leakage, and the speed is 5 drop per minute. (b) Oil conservator of #1 main transformer body has serious oil leakage. (c) The integration of (a) and (b)

### 4.2.3 Electric power knowledge graph layer

The entire electric power KG is constituted by merging a mass of sub-domain KGs towards different business scenarios or power domains in SGs. In other words, disparate domain KGs related to SGs application scenarios are integrated into a knowledge base eventually by linking correlative nodes. The electric power knowledge is updated, revised, and enriched dynamically by quality evaluation, knowledge updating and knowledge reasoning technology [100], then the updated KG will be stored in graph base, which is a dynamic and iterative process.

Taking the above two defect records as an example, oil conservator has the same noun meaning with oil storage cabinet, and the oil conservator is a component of the transformer body. The two sub-reflections are integrated after knowledge fusion, knowledge reasoning, and knowledge updating, and the relation between the oil storage cabinet and the transformer body can be clearly depicted in Figure 5(c).

From the perspective of the KG modelling in Section 3.1, a property graph like Figure 5(c) can be defined as a tuple [61], where is a set of bode ids, is a set of edges ids, is a set of labels, is a set of properties, is a set of values, maps an edge id to a pair of node ids, maps a node or edge id to a set of labels, maps a node or edge id to a set of property–value pairs.

Returning to Figure 5, the refection on knowledge graph of defect records can be depicted as follows:
    the set V contains transformer, transformer body, oil storage cabinet, and oil leakage;

    the set E contains contains and symptom;

    the set L contains power equipment, transformer component, transformer body component and transformer defect;

the set P contains ids, transformer part, body part, speed, and severity;

the set U contains 1, 1, 1, 5 drop/min, and serious;

the mapping e gives, for example, e(contains) = (Transformer, Transformer body);

the mapping l gives, for example, l(Transformer) = Power equipment and l(Transformer body) = Transformer component;

the mapping p gives, for example, p(Transformer) = (ids, 1) and p(Oil leakage) = (speed, 5 drop/min).

Until now, knowledge graph towards smart grids can be established through three above-mentioned key steps as shown in Figure 6, which consists of four vital knowledge graph/base, namely, entity knowledge graph of power equipment, concept knowledge graph, fault case knowledge base, and business logic knowledge base. The entity KG consists of all kinds of primary or secondary power equipment in SGs. Meanwhile, the voltage, power, frequency, other attributes, and connection relationships of the equipment can be updated according to the real-time operation data of the power grid. Concept KG is built by abstracting the entity KG, which is more in line with human beings. It can be used to normalise and refine the fact expression. The business logic knowledge base is the integration of fault disposal/maintenance plan, scheduling rules, operation and maintenance principles, cause analysis, disposal points and other information, in which the relevant information can be well inquired and reasoned by incorporating path algorithms. When a new accident occurs, the fault knowledge base is updated by recording and saving the related fault information. Moreover, disposal history and operational suggestions of similar cases can be pushed by calculating the similarities of the cases in conceptual layer. In addition, the decision-making strategies related to fault diagnosis and disposal are provided by the logical operation base and rule base, which are constructed manually or semi-automatically by extracting from historical defect records, and expert experience. An example of the knowledge graph towards transformer body defect is depicted in Figure 7.

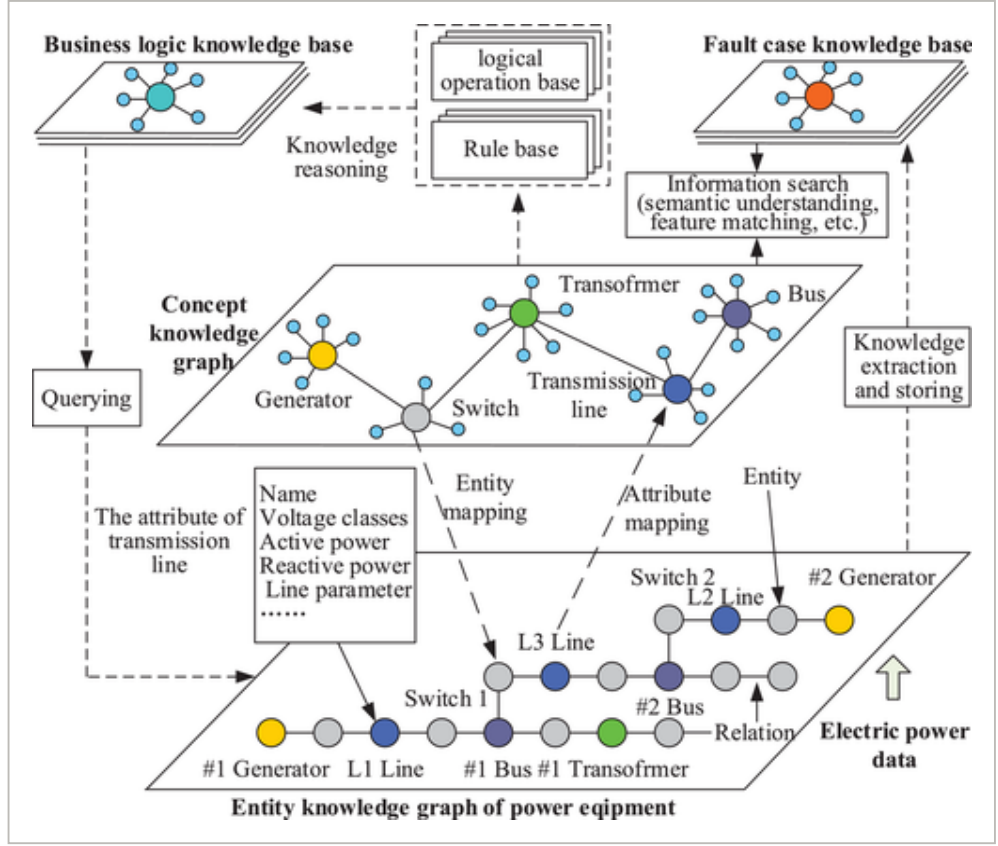

**FIGURE 6**

The integration of knowledge graph and smart grids [101]

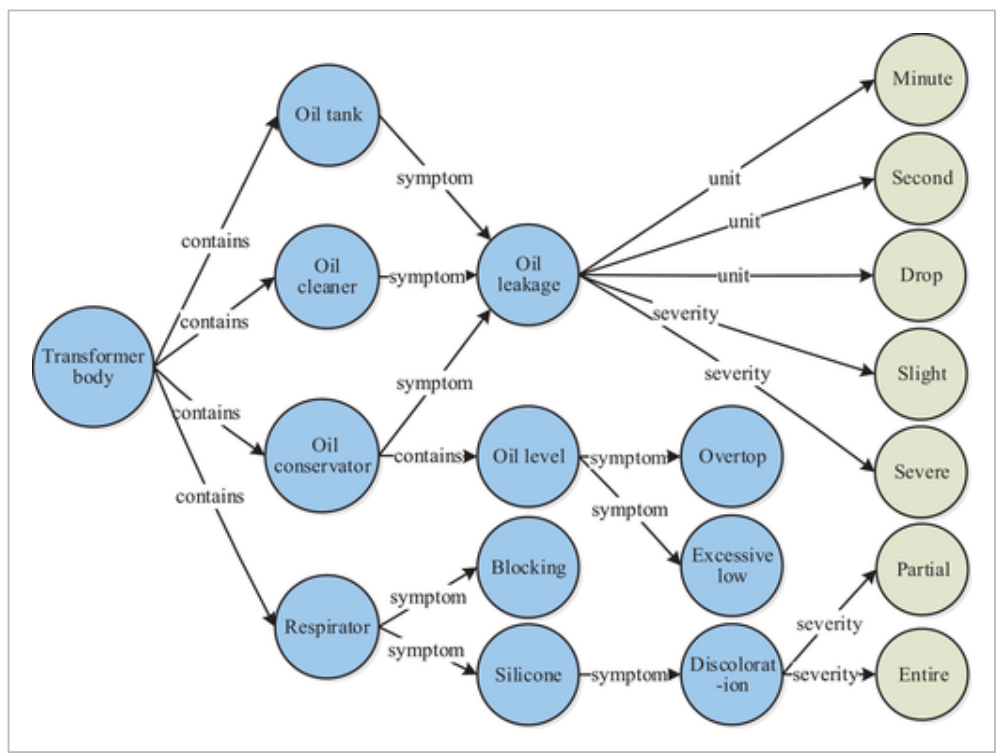

**FIGURE 7**

The knowledge graph towards transformer body defect

## 4.3 Knowledge computation and management

This part of the BKG platform towards SGs undertakes several critical tasks [92]: (a) Allocating storage space of different electric power information and intelligent function units, and updating these dynamically. (b) Taking control over the running of all functions of intelligent applications and the whole platform. (c) Taking responsibility for platform security. (d) In particular, monitoring and controlling the life cycle of the platform evolution.

### 4.3.1 Knowledge computation

The knowledge computation module manages the computing framework and algorithms, including distributed big data processing framework (Hadoop, Spark), graph computation, NLP, machine learning, deep computation and so on. Relevant electric power data is often stored in several physical repositories, but they have to be computed aiming at some dedicated sub-domain services before effective use. In particular, it is worth highlighting two kinds of data processing methods: graph computation and deep learning. In general, any knowledge structure (head entity, predicate, tail entity) of the electric power KG can be considered as a graph. Graph computation has a great advantage with regard to processing, analysing, and visualising massive data with graph structures and complicated relationships. Moreover, graph computing can explore the topologies and properties of KG4SGs via linked vertices and edges [102, 103]. To the best of our knowledge, deep learning has been widely applied in SGs in recent years [104-107]. In terms of the KGs, deep learning has been leveraged by relational reasoning [108], knowledge representation [22] and so on.

Hadoop MapReduce is deployed to process the massive electric power big data for its being good at batch processing of big knowledge computing and the superiority of distributed parallel framework. Apache Spark GraphX, a new component in Spark for graph and graph-parallel computation, is implemented in the BKG platform for graph computing. In addition to a highly flexible API (application program interface), GraphX provides users primitives for elementary graph operations. Apache Spark and Spark Streaming are used as the in-memory and streaming computing framework [109]. Since the electric power data and knowledge are mostly represented by using resource description framework (RDF), SparQL, a specialised language for RDF, is applied to perform semantic knowledge-based querying over massive information.

Above all, the BKG platform can compute massive amounts of data and knowledge, capture valuable information, and provide decision-making grounds for electric utilities, system operators, and power consumers.

### 4.3.2 Knowledge management

The most important module of knowledge management in the BKG platform is the knowledge storage module, which provides knowledge sources and allocates storage space for the electric power knowledge computation module. The BKG platform mainly manages four types of electricity knowledge: (1) RDF triplets; (2) abstract textual information; (3) images; (4) digital data [92]. Meanwhile, the knowledge storage module supports knowledge computing, graph computing, machine learning, deep learning, knowledge-based querying, etc. The general electric power sub-domain KGs are stored in graph databases, for example, HBase, Neo4j. Neo4j [110] is a high-performance NOSQL (Not Only Structured Query Language) graph database that stores structured data on a network rather than in tables.

The massive electric power knowledge is represented as SPO triples by mainstream representation technologies and ontology learning. The binary relationship between grid ontologies is integrated into the RDF graph model, hence, the knowledge base of each electric power sub-domain KG contains a huge quantity of RDF triples related to the electricity knowledge. The BKG platform mainly uses the Spark SQL (structured query language) [111] to support electric power knowledge operations for the reason that it is more efficient than Hadoop distributed file storage system [112], Hive [113], and Shark [114]. SparQL-Spark SQL query transformation is also deployed to achieve efficient RDF querying.

Another key part of knowledge management is the information security. Three groups of security techniques are ensured [92].

   Visitor: Each visitor is authorised to access the range and quantity of data and confidence levels according to his/her authorisation classes;

   Knowledge: Knowledge storage distribution is applied to store knowledge with different confidence levels in different areas of the BKG, which is protected at different security levels. In addition, the new knowledge must be carefully checked before it is integrated into the knowledge database;

   System: System security is ensured from two perspectives, namely, critical nodes encrypting and back-up.

Future, the KG is expected to incorporate with blockchain to provide more secure, synchronised, and guaranteed systems for SGs [69].

# 5 PROSPECTS OF INTELLIGENT APPLICATION

KG technology offers a vital methodology for expressing, organising, managing, and utilising massive, multi-source, heterogeneous, and dynamic data and information in an easy-to-understand, easy-to-use, and easy-to-maintain manner [108]. At present, KG technology is mainly used in recommendation systems [115], language modelling [116], question answering [117], or image classification [118], which has penetrated into the financial, medical and industrial sectors. In the medical field, KG has provided an emerging paradigm in medical information systems, including clinical decision support system [119], medical intelligent semantic search engine [120], and medical question answering system [121], making these systems more intelligent and user friendly. In SGs, we devise KG4SGs as a unified platform to represent and manage massive heterogeneous power data acquired by smart measurement and metering during the electric power production and operation at a semantic level. In this regard, the KG could be immense potential for power system smart business scenarios based on AI technologies like deep learning, NLP, machine learning. In this section, three typical application scenarios are taken as examples to expound that KG in SGs is a promising area and has real application value in terms of power customer service system, decision-making in dispatching, operation and maintenance of power equipment. In addition, applications of KG in other grid business scenarios are briefly surveyed.

## 5.1 Power customer service system

## 5.1.1 Background

A featured initiative in the future smart grids is active participation of electricity consumers in the ancillary service [122-124]. Customer value analysis can provide differentiated service for customers and implement benefit maximisation for power utilities. Liu et al. [125] proposed a quality inspection sampling algorithm using modified C4.5 algorithm to classify service calls and work sheets with or without defects. Sheng et al. [50] proposed a power customer appeals recognition model based on Adaboost, SVM and Random Forest. Lindén et al. [126] utilised historic consumption patterns to categorise electricity customers. Zhang et al. [127] studied the clustering algorithm-based electricity customers classification.

However, there are some deficiencies in power customer services in recent years: (a) the traditional power customer service methods such as 95,598 hotline telephones, business halls, and manual service make the communication costs, training fees, human resources, and other costs not affordable for most companies; (b) there also exists conditional constraints such as time (non-24-h service), venue (centralised customer service office), which hinder high-quality services; (c) given that customer service staff have disparate grasps on the business problems, the quality of service is uneven, which may let customers wait too long and even get no satisfactory answers; (d) customer service staff are prone to lose their enthusiasm in long-term response to repetitive problems; (e) if power enterprises want to build a large and sophisticated knowledge base for customer service systems, a huge amount of manpower and material resources must be expended. It is difficult to maintain and update the knowledge base later because of the complex relationship between electricity knowledge. All these problems hamper the quality and efficiency of personalised customer services.

At present, the information retrieval-based method, the most popular approach of question answering, is mainly based on keyword matching. However, this way considers only the shallow similarity of keywords related to the queried question, which overlooks the in-depth semantic information [108, 128]. SGs customer service question and answer system based on KG can map the consumer service knowledge into a knowledge base in the expression of natural language, which enhances the performance of power consumer service. Hence, it is an important issue to automatically classify knowledge, problems, and experiences in power supply services, obtain the optimal answers, and provide consumer-oriented interaction based on the massive work tickets data and the power business knowledge system for responding to the increasing number of SGs customers and the huge demand for consulting.

## 5.1.2 Intelligent customer service robot system

In order to deeply discover power customer service knowledge and improve the performance of electric power enterprise for intelligent knowledge management and application, KG technology can be applied to organise and manage information related to the power enterprises through the in-depth semantic analysis. In this regard, the architecture of the customer-centre customer service robot system based on KG, NLP, machine learning, and semantic web technology is proposed to extract the customer demands, achieve high-quality service, and improve the efficiency of the power system operation through deep semantic understanding in Figure 7. The framework not only provides 24-h online self-service interactive response service to reduce the pressure and repetitive work of online customer service staff, but also bridges the communication between power companies and customers.

The overall framework of the intelligent customer service robot system contains six key parts, including data layer, intelligent service engine, consumer service robot system, multi-channel accesses for social media, robot operation framework, and unified management platform. The core modules for each part are as follows:

> The data layer is used to extract the key semantic knowledge from domain corpora (e.g. electricity laws and regulations, common questions, electricity common sense, power thesaurus, WeChat history records, Twitter messages), and then construct an easy-to-understand semantic feature analysis model and an intelligent question answering knowledge base that can be understood by the robot. The knowledge base comprises the common knowledge base, business knowledge base, and rule base. The common knowledge base contains phrases, sensitive words, stop words, sensitive words, and common parts of speech involved in customer questions. The business knowledge base contains customer service business questions, power thesaurus, and customer service knowledge base. As for the rule base, special instructions are set in it to do some special scene tasks, and robots are enabled to make quick answers to different consulting scenarios by semantic understanding, KG matching, and knowledge reasoning.
>
> Intelligent service engine supports various engines and function modules for other parts of intelligent customer service robot system, including sentence segment, semantic analysis, chat conversation, answer processing, knowledge search management, etc. In addition, an intelligent routing distribution mechanism is implemented into intelligent customer service robot system, which can intelligently identify user priorities, reasonably allocate customer service, and improve the efficiency of manual customer service in terms of user membership levels, source channels, demand categories, key behaviours, business nodes, customer service division, response time, user satisfaction, conversion rate, and workload, etc.
>
> Consumer service robot system can accurately and quickly identify and predict customer intentions, cover various scenarios, and promote non-blocking human–machine communication. Moreover, the intelligent customer service interaction system can record the content that cannot be answered and classified by the current KG, which can help add the new user concerns to the existing knowledge map by manual intervention later. Proactive perception and active services are developed for continuously improving service quality, such as reminding the users whether they need to query the content that they often pay attention to, pushing the attention information for the user and so on.
>
> Multi-channel accesses for social media can receive massive customer consultation data from various channels such as chat sessions, online messages, customer evaluations from Twitter, WeChat, APP, websites. Most common problems can be replied automatically by the consumer service robot with semantic understanding.
>
> Robot operation framework supports communication with power consumers based on interactive business logic and provides service interface for the consumer service robot system, multi-channel accesses for social media, and unified management platform. This working framework can be further improved in the future.

Unified management platform implements robot management, marketing management, knowledge management, authority management, and user management, which is the basis of the whole intelligent consumer service robot system.

The intelligent customer service robot system based on KG has the following advantages, which are therefore suitable for intelligent customer service in the era of the AI:

Reducing customer service costs;

Achieving high-quality and more timely responses;

Fulfilling strong customer-centre expectations;

Improving the explainability of the answers.

Until now, there have been several KGs and research focusing on intelligent question answering engines [136-139]. However, there are few explorations on KG applied to intelligent customer service towards SGs. Tan et al. [140] proposed a hybrid domain features' KG smart question answering system to reduce the ambiguity of Chinese language questions and the cost of online service operation and maintenance. The entity was identified by long short-term memory model and the semantic enhancement method based on the topic comparison was proposed to find external knowledge. However, the answer was obtained by using heuristic rules.

At present, there appears an increasing number of demands for high-quality and reliable electric power and services. However, large-scale incorporation of distributed renewable energy permeates into the power systems, affecting the stability and quality of power production to a certain extent. The involvement of power consumers may promote the interaction and responsiveness of the customers in different ways via emerging handsets and other connected devices, which will transform the enterprise–customer relationship and make power customers as participants [141, 142]. In particular, with the advent of mobile social media era, mobile applications such as Twitter, WeChat provide consumers with more chances to get involved in electricity services to resolve their requests more efficiently and conveniently. In the future, power utilities will obtain the long-term trust of customers as power service advisors. In particular, intelligent customer service robot system can fulfil strong customer-centre expectations, such as personalised services, proactive electrical energy saving tips, improving customer engagement, and expanding power customer experience initiatives in a user-friendly way.

## 5.2 Decision-making in dispatching

### 5.2.1 Background

In the existing dispatching mechanism, operation rules are made offline by experts and the "empirical + analytical" model is still taking dominant position in dispatching business processing. However, this cannot satisfy the requirements for online application with power system enlarged, the tightness of network interlinks, and enhanced complexity degree of system. There are some challenges in SGs operation:

(1) Problems in the formulation of operating rules. The operating rules are mainly based on offline analysis and manual induction, and the results depend on developed expert experience. This method is time-consuming and labour intensive, and can only contrapose the typical operation modes, which cannot adapt to all online situations. With the expansion of power grids and the complexity of network structure, the acquirement of operating rules becomes more difficult.

(2) Problems in the application of operating rules. The obtained rules are usually unchanged for a long time because of the huge workload of formulating rules, which causes great difficulty in matching the online operation modes. As a result, electric power dispatcher has to adopt conservative operating limit values, corresponding with rough rules and poor economics. On the other hand, the rules are optimistic for not considering a few extreme operating modes and accidents in offline analysis, which may threaten power system security.

(3) In the large-scale incorporation of intermittent renewable energy, frequent occurrence of natural disasters caused by climate change, and increasingly time variation and complexity of the power grids operation modes, traditional scheduling decision-making mechanisms can no longer meet the online operation, which has posed several challenges to the operation of complex power grids.

With the development of AI technology, numerous research on decision-making in dispatching has been carried out. Genc et al. [143] applied decision tree to develop the preventive and corrective controls. Zhu et al. [144] utilised the imbalance learning to assess power system dynamic stability. Zheng et al. [145] developed a multi-objective group search optimiser with adaptive covariance and Lévy flights to optimise the power dispatch of the large-scale integrated energy system. Liu et al. [146] proposed an energy network dispatch optimisation under emergency of local energy shortage with web tool for automatic large group decision-making. Li et al. [147] studied a two-stage methodology by combining multi-objective optimisation with integrated decision-making. However, large-scale incorporation of intermittent renewable energy has posed great threats to the economic dispatch of smart grids [148-151]. To address the high levels of uncertainty associated with the intermittency of resources [152], one issue is to break through the computational complexity. This may be solved by the implemented forecasting system, such as load and renewable resources generation forecasting [153-155]. Sparse Bayesian classification and Dempster–Shafer theory-based wind generation [156], Markov chain-based stochastic optimisation of economic dispatch [157], spatio-temporal wind power forecasting [158], fuzzy prediction interval model-based renewable resources and loads forecasting [152], and wind and locational marginal price forecasting-based dispatch scheduling [159] are studied detailedly, which lead to another problem. That is, several studies are based on black-box models, which are less interpretive and difficult to be accepted by field operators.

On the other hand, there are a great deal of texts related to scheduled routine operations, which make fault handling for power grids dispatching, knowledge and experience learning based on unstructured Chinese language become obstacles for the collaboration of machine and human being. Therefore, it is imperative to establish automatic operator (AO) for decision-making in dispatching by utilising AI and NLP to extract knowledge from long-term operating experience, dispatching rules, and tickets for the demand of accommodating the SGs dispatching business.

### 5.2.2 Automatic operator for decision-making in dispatching

The AO for decision-making in dispatching equips the machine with cognitive capabilities and dispatching knowledge engines are implemented by knowledge retrieval, knowledge fusion, and knowledge reasoning, thus improving the automation and intelligence of dispatching. The overall framework (Figure 8) comprises four key steps: establishing a corpus and semantic model of dispatching professional lexicons targeting at the characteristics of text terms and complex scene distribution in SGs dispatching; extracting information from dispatching rules and tickets by NLP to form a computer recognised machine language and deeply digging the critical interfaces of SGs, key factors affecting limit transmission capacity of interfaces and their quantitative relationships [160]; learning textual fault disposal rules automatically; and finally building a large-scale knowledge base of dispatching rules and fault processing schemes for supporting AO decision-making. Instead of rough and offline dispatching rules, fine rules can be acquired online by AO. As a result, AO promotes intelligent decision-making in dispatching and automatic accident disposal. When a dispatcher interacts with AO, AO can identify the dispatcher intention, capture the critical information by entity recognition, syntactic analysis, and semantic analysis, match and retrieve the processing results from the KG of dispatching rules and fault processing schemes, and then return the disposal strategy or accident warning of requests. In this way, it not only expands the recommended results and improves the accuracy, but also enhances human–computer interaction services [108].

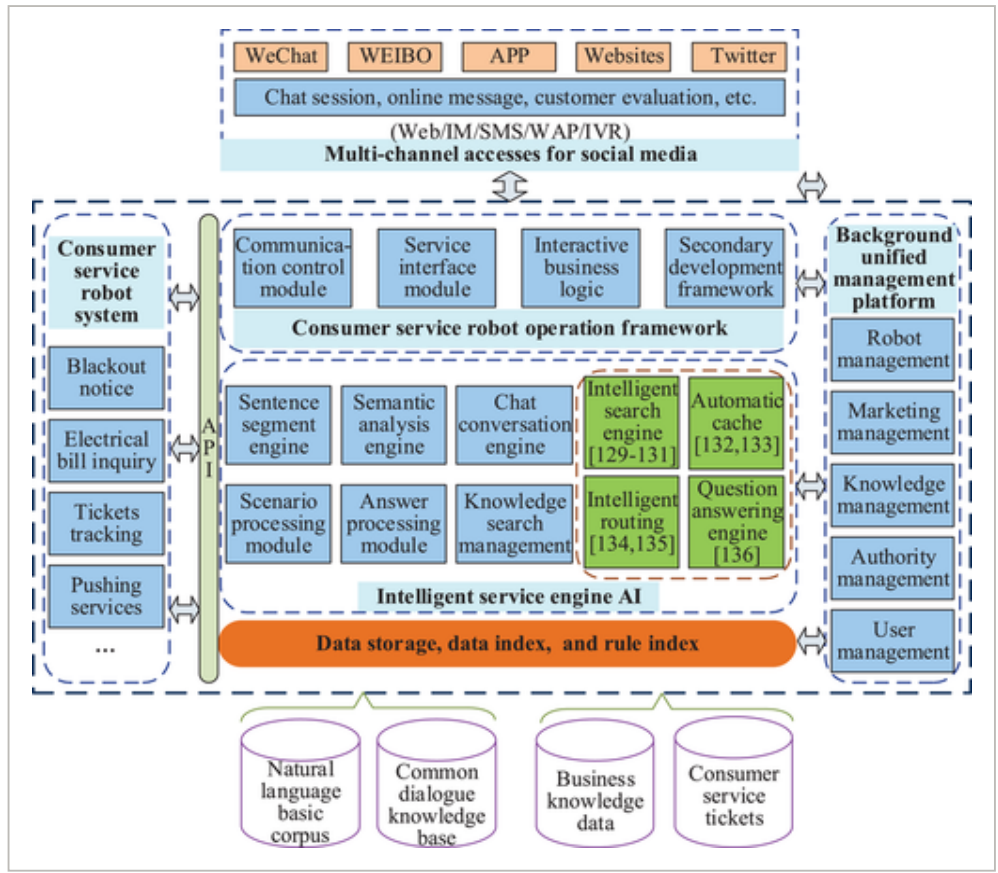

**FIGURE 8**

Open in figure viewer | PowerPoint

Intelligent customer service robot system architecture towards SGs

The AO for decision-making based on the KG has the following advantages:

Extracting the relationship between the operation modes and the power flow interfaces from the textual tickets of SGs operation modes by NLP technology;

Updating the interface stability limits automatically in accordance with the real-time operation state of SGs, which avoids the situation in which the interface limits are not updated timely due to slow manual judgment;

- Capturing the key parameters that affect the limit transmission capacity of the power flow interfaces and its quantitative relationship, which has strong interpretability and high accuracy;
- Providing real-time decision-making information for interface control, improving the interface transfer capacity, and overcoming the shortcoming of the poor interpretability of traditional black-box models to some extent.

Until now, there are few pieces of research and achievements of KG construction in power grids dispatching. Li et al. combined 'top-down' with 'bottom-up' to construct a KG of the power dispatching automation system [161]. The constructed KG of remote measurement helped understand the business relationship of the whole system and facilitated fault analysis when the system fault occurred, which proved that KG could be well applied in the intelligent auxiliary decision-making. Shan et al. introduced the key technologies and technical routes of intelligent assisted decision-making technology based on KG. It proved that the KG technique can be applied to the inference and analysis of dispatching rule knowledge by taking the fault disposal technology as an example [162]. State Grid Corporation of Hangzhou designed and deployed an AI-based virtual dispatching assistant named 'Pach', which realised fault judgment, plan issuing, and repair command via more than 5000 h of speech training and a large number of safety regulations, work cases and professional papers learning based on KG [163].

As a knowledge map of the industry, the KG for SGs dispatching has its unique professional characteristics and the accuracy can only be improved by integrating comprehensive power knowledge. In the future, intelligent decision-making for SGs dispatching tends to be based on the sub-domain KG (e.g. graphs for power equipment, concepts, operation rules and fault cases) and electric power field professional lexicons. And quicker disposal strategies will be obtained from AO, which will further help dispatchers capture the current status and development trends of SGs actively, quickly, comprehensively and accurately. The AO for decision-making will learn automatically and KG will be promoted iteratively corresponding with the continuous changes of the SGs. Moreover, AO can reduce the risk of manual handling errors and expend intelligent applications in power grids dispatching.

## 5.3 Operation and maintenance of power equipment

### 5.3.1 Background

In SGs, electric power equipment is playing an important role in the generation, transformation, transmission, and distribution of electricity energy. The enormous investment and increasing demands for electricity energy motivate utilities to accurately assess and diagnose the condition of power equipment assets [164]. As a result, condition-based maintenance (CBM) comes into being. CBM can alter scheduled maintenance, prolong the service life of power equipment, and save the time and cost of maintenance, thus making maintenance work more scientific. In general, CBM decision-making is gradually applied in SGs [165].

At present, CBM for power transmission and transformation equipment analyse and judge the health condition mainly through one or a few state parameters and unified diagnostic criteria, which fails to comprehensively take full advantage of defects, maintenance history, family quality history, etc. This work procedure is difficult to meet the demands of differentiated, meticulous, and personalised condition assessment and fault diagnosis. As a result, there are inevitable excessive maintenance and insufficient maintenance, which may lead to enormous waste of manpower and resources. In short, existing research on operation and maintenance of power transmission and transformation equipment are not holistic, systematic, or optimal, whose defects are listed as follows:

(1) The informatization of operation and maintenance is quite poor. Although massive equipment status information that scatters in various departments of the power systems is collected, multiple heterogeneous data cannot be managed uniformly due to the lack of unified protocols and standards;
(2) Most of the existing equipment evaluation and diagnosis technologies are mainly based on a single or a few monitoring parameters, which cannot comprehensively reflect the various operation information of different equipment, resulting in poor results of diagnosis and evaluation;
(3) The current methods of condition assessment and alarming still rely on experts' operating experience to a large extent, and the theories and methods of equipment fault feature selection and condition assessment models cannot meet the requirements of SGs scenarios;
(4) Intelligent decision-making and management have not yet been coming into being.

The fast developments and applications of AI provide new opportunities for the operation and maintenance of power equipment, such as defect recognition, prediction, and fault diagnosis. For example, regression prediction [166], SVM [167], artificial neural network [168, 169], grey model [170], and combinational model [171, 172] have been widely applied in operating condition prediction of power transformer.

As for operation and maintenance of power equipment, advanced measurement infrastructures have been deployed in smart grids, and equipment state data has gradually emerged as large-volume, multi-type, and fast-growing, thus paving the way for the application and development of big data, artificial intelligence, and other technologies in operation and maintenance. Hence, integrating big data and knowledge is valuable for differentiated and comprehensive CBM.

Since massive heterogeneous data and information exist inside and outside the plants, it is of great importance and necessity to exploit new techniques and maintenance engineering to build a unified understanding of health condition and fault features, develop multi-dimensional information fusion technology, conduct personalised condition assessment, set up decision-making strategies for maintenance, and support intelligent operation and maintenance for electrical power assets managers.

### 5.3.2 Intelligent operation and maintenance based on multi-modal knowledge graph

Given that there are massive heterogeneous data collected by the AMI and existing research cannot integrate the comprehensive information of power equipment, a framework is proposed to establish a multi-modal KG as shown in Figure 9, which integrates structured data, images, Wikimedia, defect records, and maintenance tickets. This model supports personalised condition assessment and diagnosis, and CBM decision-making strategies towards the operation and maintenance of power equipment. In text processing flow, the textual defect records and maintenance tickets are processed by knowledge extraction and semantic analysis in view of text context content mining methods and numerous entities, relations, and attributes related to transmission and transformation equipment are obtained. In visual processing flow, visual semantic information of each image of power equipment is acquired by a classic deep neural network (DNN) model and corrected by human intervention. In the process of constructing the cross-modal KG, the key is to establish a unified knowledge description model of heterogeneous data, blend the visual semantic information, structured data, and context semantic information of text content and discover the comprehensive and effective extension concepts, as well as extract concept relationship and semantic hierarchy. Considering the emerging visualisation applications based on cross-modal retrieval with the image-text-structured-data knowledge, linked multi-modal KG can be widely applied to the operation and maintenance of power equipment, including personalised condition assessment, diagnosis, and CBM decision-making strategies.

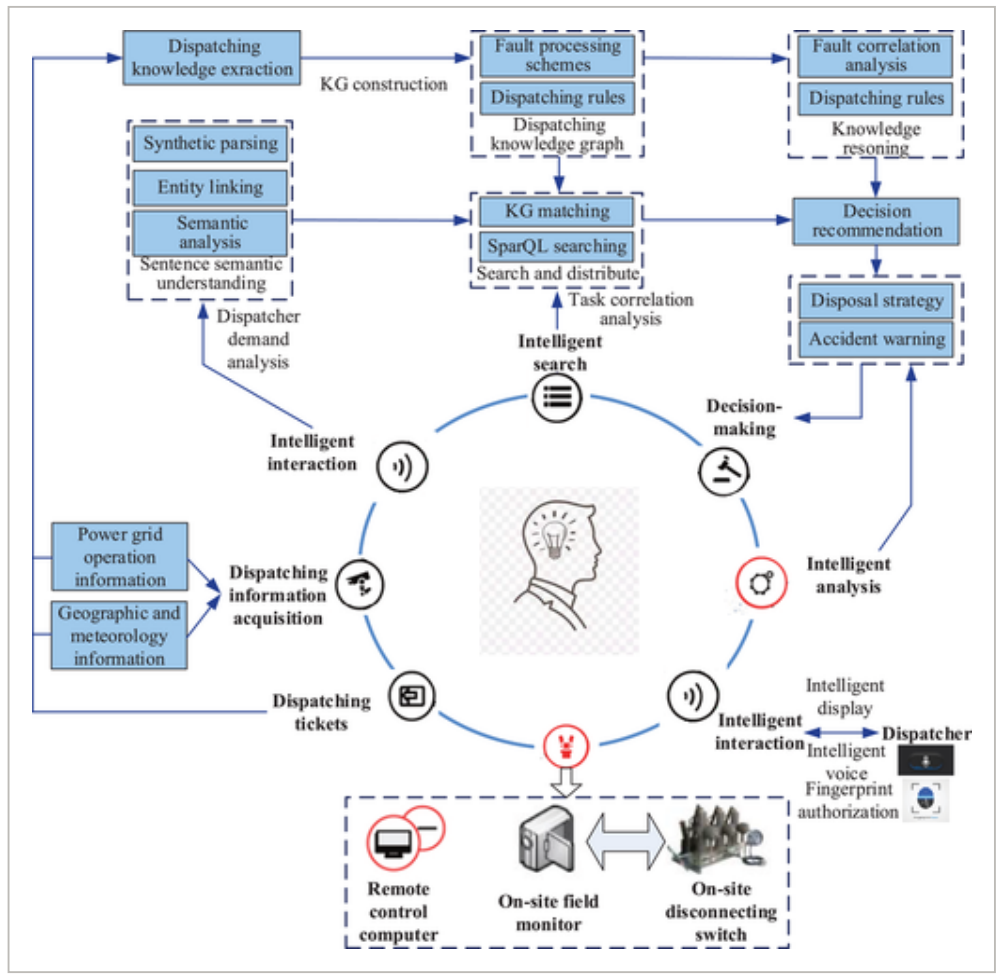

**FIGURE 9**

Open in figure viewer | PowerPoint

Automatic operator for decision-making in dispatching based on knowledge graph

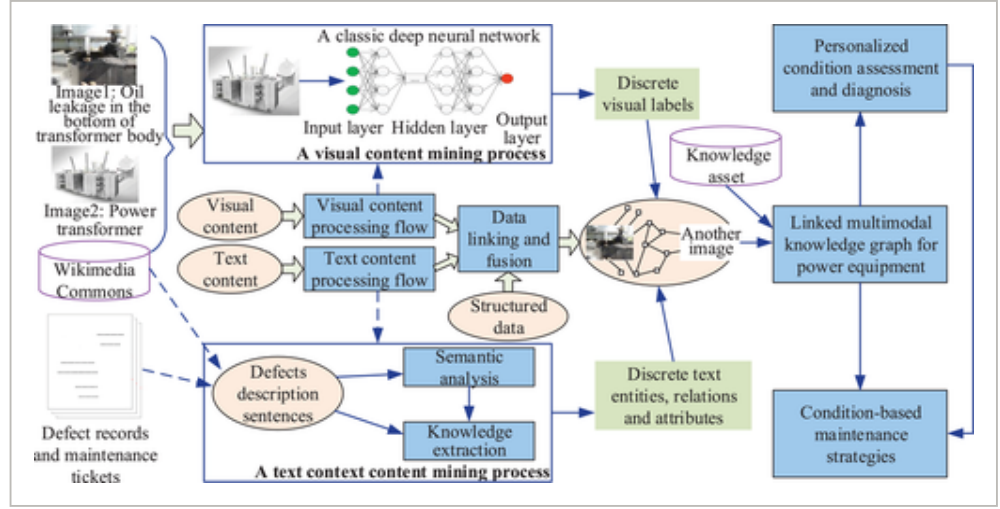

**FIGURE 10**

Open in figure viewer | PowerPoint

A framework of multi-modal knowledge graph for power equipment

In contrast with existing methods, intelligent operation and maintenance based on multi-modal KG for power equipment have four advantages:

Merging data of transformer equipment from the data layer perspective. Existing information fusion of power equipment operating parameters is mainly concentrated on feature layer fusion and decision layer fusion;

Integrating with the early-developed expertise for the condition assessment, diagnosis, and maintenance in the long-term electric power production, transformation, transmission, and distribution;

Achieving differentiated, refined and comprehensive assessment results by combining the data from different sources such as power grids operation, equipment status, meteorological environment, and analysing the current and historical state changes of the power equipment;

Providing the theoretical basis for equipment operation and maintenance decision-making. Over the last few years, KG relevant techniques have been extensively used to develop more accurate condition assessment and diagnostic tools. Liu et al. adapted KG to retrieve defect records of power equipment, which could significantly improve the retrieval effect of defect records [69]. Wang and Liu [173] utilised KG technology to construct an error recognition model of power equipment defect records. Moreover, Tang et al. [11] applied KG into merging heterogeneous data of all the power equipment to enhance the management of power equipment, which could query the basic information, classify relevant information of products, maintain the integrity of information, demonstrate the relationship between products, and achieve real-time updates on the content of any product. Cui [174] introduced KG to integrate the data of the IoT system for detecting the abnormal state of electric device. In addition, Su et al. utilised the KG to integrate multi-source information from power terminal equipment in the Ubiquitous Power Internet of Things based on relational databases and ontologies [175]. In general, the KG technique has been applied preliminary in the operation and maintenance of power equipment. The research on the multi-modal KG for intelligent operation and maintenance of power equipment should be heightened in the future, such as multi-modal attribute expression, complex multi-modal relationship mining, unified representation, and incremental updating of multi-modal KG.

## 5.4 Other applications

There is several research on the studies of KG applied in other power fields, such as low-voltage distribution network topology verification [176], visual query method for large blackouts [177], and the generation of secondary security ticketing [178]. In [176], data in multiple low-voltage distribution network information systems was integrated, and the KG for low-voltage distribution network topology was built. The household transformer relationship could be well verified and identified by the constructed KG. In order to analyse the causes of large blackouts based on large volume of Chinese text, Zhang et al. [177] utilised the web crawler, deep learning, and knowledge graph technology to capture the event entity, relation and attribute. Moreover, a large blackouts knowledge graph was built and visual queries of nodes, relations, and paths for large blackouts were implemented. In terms of the generation of secondary security ticketing, Wang et al. [178] adapted the KG to construct the search engines of intelligent substations, which provided the unified data integration and enhance operational efficiency.

# 6 ISSUES AND CHALLENGES

As we mentioned in Section 5, KG is a promising and fast-developing research topic in the customer service robot system, power semantic search engines, question answering engines, decision-making, automatic operator, and other applications towards SGs. The combination with better KG and SGs would offer new opportunities for AI service providers, which in turn presents many challenges. To move on an important step towards KG4SGs, there is a need for a better understanding of the issues and challenges in the development of electric power KG and its technologies, which are as follows.

## 6.1 Making use of heterogeneous electricity knowledge

With rapid development of SGs technology and wide deployment of the measurement devices, an unprecedented amount of electric power big data has been obtained. Electric power big data is characterised by various sources (e.g. PMS, EMS, GIS), high volume (thousands of terrabytes), wide varieties (e.g. digital data, textual defect records, images, social media), varying velocity (e.g. online monitoring, daily inspections, quarterly/yearly maintenance), veracity (e.g. missing data, redundancies, malicious information), and values (e.g. operational, technical, economic) [176, 177]. How to efficiently and effectively utilise these multi-source heterogeneous information is becoming a critical and challenging problem [70]. However, there have been no unified data representation and logical structure for the fragment electricity data and knowledge so far. Moreover, complex power network structure and diversification of access to information lead to increasingly prominent data heterogeneity and 'information island'. Furthermore, the relationships of multi-source heterogeneous data become more complex and evolve over time. To extract valuable knowledge from massive heterogeneous power data depending only on the domain knowledge of traditional expert systems is neither efficient nor sufficient. Hence, it is essential and beneficial for all the stakeholders in their power sector to develop more efficient models to better make use of multi-modal electricity knowledge and unlock underlying information and relations by the cross-fertilization of the multi-source power data [181].

## 6.2 Constructing dynamic professional lexicons in the electric power field

The construction of general KG can be improved by the massive semantic information in semantic web, in which all the data is formal, structured, and shareable. However, as for general domain-specific KG, merely few formal, structured, and shareable information could be obtained on the Internet, especially the definite closed electric power field. All needed data and information are closely linked to SGs, which poses great challenges to the knowledge mining and the construction of electric power KG. The construction of professional lexicons in the electric power field could improve this issue to some extent. The quality and quantity of professional lexicons not only determine the accuracy of word segmentation and sentence context comprehension in text pre-processing, but also affect the performance of eliminating ambiguity and constructing KGs. Meanwhile, professional lexicons in the electric power field are not static and change over time due to electric power big data characterised by massive, heterogeneous, and multiple sources [62, 182].

On the other hand, there are many sub-domains in SGs, each of which has a different requirement for professional lexicons of the electric power knowledge. For example, GIS refers to gas-insulated switchgear in power systems. However, GIS is geographic information system from the perspective of computer science and technology. In addition, the emergence of new concepts such as the Energy Internet, integrated energy services, and ubiquitous power IoT, has produced plenty of new vocabularies, including a great number of highly professional industry words. With the development of SGs businesses, more and more new concepts and vocabularies will continue to emerge, and traditional knowledge mining methods can no longer adapt to these challenges. Therefore, it is of great significance to construct dynamic high-quality lexicons to improve the accuracy of text mining, entity extraction, and knowledge processing in this under-explored domain.

## 6.3 Improving the quality of KGs

One of the main challenges in electric power KG applications is the quality of large-scale KGs themselves. Existing research on measuring the integrity and legitimacy of the generated KGs cannot meet the demands for SGs scenario applications due to the lack of an effective electric terminology validation and verification model. Thus, newly constructed KGs for SGs inevitably suffer from noises, conflicts, and incompleteness. The incorrect and missing knowledge lead to error propagation in SG scenario applications. In this regard, the issues of knowledge fusion, knowledge reasoning and quality evaluation calls for future research. Specifically, electricity facts from different power data sources and sub-systems have to be carefully checked and those with high levels of 'confidence' can be integrated an unified knowledge base. Then the constructed KG should be refined by add missing knowledge and identifying and removing errors [25].

In terms of knowledge quality assessment, it is the future research goal of this field to build a perfect quality assessment technical standard or index system. Hence, it is urgent to establish the evaluation criteria to quantify the credibility of domain knowledge for achieving accurate construction of huge KGs for SGs, automatically detect the conflicts and errors and add missing knowledge [22]. The crowdsourcing techniques have already been applied in entity linking and entity resolution, and human–computer cooperative crowdsourcing algorithms can improve the quality of knowledge fusion. The design of the crowdsourcing algorithm requires a trade-off among the amount of data, the quality of knowledge base alignment and manual annotations. It is expected to make influencing breakthroughs in combining the crowdsourcing platform with the knowledge base alignment model organically and effectively judging the quality of other workers' annotations [183-185].

## 6.4 Multilingual KGs of different sub-domains

An entire electric power KG contains a vast mount of power facts, events, and entities excavated from different electric power scenarios. Each electric power domain-topic KG contains a huge amount of knowledge featuring complicated structure and data, which pose certain challenges to the accuracy and efficiency of the knowledge fusion (e.g. entity link, entity alignment, entity resolution) and knowledge reasoning. In this regard, future research should focus on merging knowledge of different electric power sub-domain KGs. Moreover, multi-language KGs corresponding with different SGs application scenarios will be constructed eventually in the future [186], which are closely associated with the trait of natural language and electricity knowledge. The complementary capabilities of multilingual knowledge bases provide more possibilities for real applications of KGs, presumably power consumer service system, automatic operator for decision-making in dispatching, power knowledge intelligent search, and smart question answering system [187].

## 6.5 Knowledge updating of large-scale KGs for SGs

Logically, the update of knowledge bases for SGs consists of the concept modelling layer and the data layer. In the concept layer, novel and complex concepts related to electric power knowledge are abstracted and decomposed. This layer assists in understanding, managing, and constructing knowledge base technologies aiming at the semantic description of human–computer interaction. Obviously, the updating of concept might affect all direct or indirect sub-concepts, entities, and properties. The main objective of the data layer is to add, modify, and delete entities, relationships, and attribute values related to electricity knowledge. In the data layer, multiple factors such as the reliability, uncertainty, consistency (i.e. conflicts or redundancy) of the electric power data and knowledge should be involved and considered comprehensively.

Existing knowledge updating technology relies heavily on manual intervention, which is time-consuming and labour intensive [188]. This means that existing methods may lead to difficulties in full utilisation and further development in actual SGs scenarios due to the limitations of their models and computational complexity. Hence, it is crucial to design a novel framework of knowledge updating which can carry out online learning and update new electricity knowledge incrementally and automatically for various applications of KGs in SGs. Incremental updating technology [189, 190] is the future research hot spot in the field of knowledge updating, which utilises the existing knowledge to achieve rapid updating of knowledge and consumes fewer resources. Moreover, how to ensure the effectiveness and efficiency of automatic updating is another major challenge.

# 7 CONCLUSIONS AND FUTURE DIRECTIONS

Over the past few years, the SGs have grown from an idea into a world-wide recognised topic. The gathered massive data contain much knowledge, and much of electric power knowledge cannot be expressed, managed and analysed sufficiently and effectively, which is rarely applied in power grids scenarios. However, the KGs provide a feasible and practical means to combine the electric power big data processing with robust semantic technologies, making first steps towards the new generation of artificial intelligence in SGs. Moreover, KGs offer the unified knowledge representation way to achieve the reflection of electric power concept and facts, events, and entities of power systems by ontology learning, professional power vocabularies, etc. Hence, operational activities, such as querying, decision-making, appeals analysis, and high-quality personalised services can be performed by KG4SGs. It is increasingly significant to build an integrated platform for electric power knowledge acquisition, mining, representation, fusion, and SGs business applications.

In this paper, we have reviewed the current techniques of knowledge mining, particularly the typical methods of electric power information leveraged by KGs. After that, the definition and advantages of KG and the motivation of KG applied in SGs are discussed. Then, a BKG platform towards SGs which discovers, integrates, manages, and analyses the massive electric power knowledge from a semantic perspective is proposed. Specifically, the overall framework, specific module design, and some more advanced techniques of the BKG platform towards SGs are elaborated on. As for electric power KG construction, three layers and related technologies are presented in detail, that is, electric power knowledge acquisition, sub-domain KGs of different business scenarios or power domains, and the entire electric power KG. Finally, this paper explores three typical applications of KG in SGs scenarios, namely, intelligent customer service robot system, automatic operator for decision-making in dispatching, intelligent operation and maintenance of equipment based on multi-modal KG.

The proposed BKG platform based on KG improves the knowledge expressing, understanding and sharing and supports the SGs scenario applications for knowledge retrieval, decision recommendation, and data visualisation aiming at business services. It is of significance to verify the proposed KG platform in the future. The combination of the KG technique, NLP and electric power knowledge will promote the intellectualization of the power networks. Although many basic attempts have been made to obtain the electric knowledge and manage operation and maintenance expertise, they were neither perfect nor in-depth. With increasingly complicated power systems and the continuous incorporation of intermittent renewable energy penetration, the exploration, expansion and automation of KG4SGs might lead to fruitful results in this field. In the future, the control and operation strategies of the power grids will be automatically obtained and recommended by KG4SGs according to the panoramic perception of environment and state information, supporting the autonomous and automatic operation of the power system network.

There are several ongoing or future research directions of utilising KG in SGs, which are follows.
1) **Combination of model, data, and knowledge**. Operation and control of the power system depend on the physical mechanism and modelling analysis, however, knowledge map technology belongs to the knowledge analysis method, which brings accessible, explainable, and structured information from the semantic perspective. In addition, a great deal data driven-based machine learning algorithms have been well applied in SGs for its superiority of mining vital information from massive data, such as load and new energy resources forecasting, equipment fault diagnosis. The performance of machine learning may be improved by incorporating the additional knowledge into the training process. Therefore, in the actual fault handling or dispatch, it should be combined with physical mechanism-based modelling analysis, machine learning-based recognition, prediction and fault diagnosis, KG technology and expert knowledge and experience, so as to capture underlying, ambiguous, and complex relations between entities hidden in electric power big data and realise the comprehensive analysis and treatment of grid fault.

2) **New techniques for electricity knowledge**. A majority of existing research on knowledge extraction, knowledge representation, and knowledge reasoning is based on web data. However, when it comes to electric power systems, there are relatively sparse web documents related to SGs scenarios, and pattern-based methods cannot directly be applied in SGs. Moreover, many rules and experiences are hidden in the extensive professional background knowledge, which leads to great difficulties in knowledge extraction and mining for a electric power domain. Hence, how to extract entities and relations related to professional electricity knowledge accurately is particularly crucial [186]. Furthermore, how to effectively conduct the knowledge fusion of manually defined knowledge and knowledge obtained by machine automatic learning in SGs is of great research value and difficulty, especially in terms of the unified representation of knowledge, the resolution of knowledge conflicts, the updating of knowledge, etc. In the future, it is dispensable to construct the large-scale professional lexicons to provide more semantic correlation information and improve the accuracy of knowledge extraction, knowledge representation, and knowledge reasoning towards SGs. Meanwhile, an important step is to study new techniques for electricity knowledge based on existing KG technologies to tackle these issues.

3) **KGs towards SG scenarios**. Since big data in SGs is characterised by big volume, wide varieties, varying velocity, veracity, and values [177, 191], this task will encounter several difficulties: the information and knowledge related to SGs scenarios are characterised by diversity, correlation, synergy, and hiddenness; owing to the unique characteristics and complexities of different SG scenarios, it is still difficult to effectively and efficiently apply KG technologies in SGs, and; the existing KG technologies are too cumbersome to deploy in the actual SGs scenario applications [22]. Therefore, in terms of different business demands for power systems, future work should focus on both effective and efficient methods to implement and establish large-scale sub-domain KGs towards SGs scenarios.

4) **Interpretability of existing models**. Give that most machine learning models applied in SGs (especially DNNs) are typical black-box models, and the algorithms cannot give convincing explanations for their results. This issue may be explored by KG, which is an AI technology with rich semantic and logical expression ability. Once the electric power KG is established, all electricity knowledge is heavily interconnected by the nodes and edges. Machine learning models on graph allow improvisation in the interpretability of transforming computer-aided experiments into decision-making analysis to some extent [62].